\pdfoutput=1

\documentclass[11pt]{article}

\usepackage{EMNLP2022}

\usepackage{times}
\usepackage{latexsym}

\usepackage[T1]{fontenc}

\usepackage[utf8]{inputenc}

\usepackage{microtype}

\usepackage{inconsolata}

\usepackage{graphicx}
\usepackage{url}
\usepackage{latexsym}
\usepackage{tabularx}
\usepackage{adjustbox}
\usepackage{multirow}
\usepackage{booktabs}
\usepackage{xcolor}
\usepackage{amsmath}
\usepackage{makecell}
\usepackage{hyperref}
\usepackage{nameref}
\usepackage{makecell}
\usepackage{amssymb}
\usepackage{pifont}
\usepackage{enumitem}

\newcommand\wecEng{WEC-Eng}
\newcommand\weces{CoreSearch}

\newcommand{\cmark}{\ding{51}}%
\newcommand{\xmark}{\ding{55}}%

\definecolor{greenrgb}{RGB}{34,139,34}
\definecolor{lightblue1}{rgb}{0.6, 0.81, 0.93}
\definecolor{lightblue2}{rgb}{0.45, 0.76, 0.98}
\definecolor{lightblue3}{rgb}{0.35, 0.76, 0.98}

\title{Cross-document Event Coreference Search: Task, Dataset and Modeling}

\author{Alon Eirew\textsuperscript{1,2} \quad
        Avi Caciularu\textsuperscript{1} \quad
    Ido Dagan\textsuperscript{1} \\
\textsuperscript{1}Bar Ilan University, Ramat-Gan, Israel \quad \textsuperscript{2}Intel Labs, Israel \\
  {\tt  alon.eirew@intel.com} \quad \\ {\tt  avi.c33@gmail.com} \quad \\ {\tt  dagan@cs.biu.ac.il} \\
  }

\begin{document}
\maketitle
\begin{abstract} 
The task of Cross-document Coreference Resolution has been traditionally formulated as requiring to identify \textit{all} coreference links across a given set of documents. We propose an appealing, and often more applicable, complementary set up for the task -- \textit{Cross-document Coreference Search}, focusing in this paper on event coreference. Concretely, given a mention in context of an event of interest, considered as a query, the task is to find all coreferring mentions for the query event in a large document collection. To support research on this task, we create a corresponding dataset, which is derived from Wikipedia while leveraging annotations in the available Wikipedia Event Coreference dataset (\wecEng{}). Observing that the coreference search setup is largely analogous to the setting of Open Domain Question Answering, we adapt the prominent Deep Passage Retrieval (DPR) model to our setting, as an appealing baseline. Finally, we present a novel model that integrates a powerful coreference scoring scheme into the DPR architecture, yielding improved performance.





\end{abstract}

\section{Introduction}
\label{sec:intro}
Cross-Document Event Coreference (CDEC) resolution is the task of identifying clusters of text mentions, across multiple texts, that refer to the same event. For example, consider the following two underlined event mentions from the WEC-Eng CDEC dataset \citep{eirew-etal-2021-wec}:

\vspace{-5pt}
\begin{enumerate}[leftmargin=*]
    \item \emph{...On \textbf{14 April 2010}, an \underline{earthquake} struck the prefecture, registering a magnitude of 6.9 (USGS, EMSC) or 7.1 (Xinhua). It originated in the \textbf{Yushu Tibetan} Autonomous Prefecture...}
    \item \emph{...a school mostly for \textbf{Tibetan} orphans in Chindu County, Qinghai, after the \textbf{2010 Yushu} \underline{earthquake} destroyed the old school...}
\end{enumerate}
\vspace{-5pt}

Both event mentions refer to the same earthquake, as can be determined by the shared event arguments (2010, Yushu, Tibetan). In event coreference resolution, the goal is to cluster event mentions that refer to the same event, whether within a single document or across a document collection.

Currently, with the growing number of documents describing real-world events and event-oriented information, the need for efficient methods for accessing such information is apparent. Successful and efficient identification, clustering, and access to event-related information, may be beneficial for a broad range of applications at the multi-text level, that need to match and integrate information across documents, such as multi-document summarization \citep{falke-etal-2017-concept,liao-etal-2018-abstract}, multi-hop question answering \citep{dhingra-etal-2018-neural, wang-etal-2019-multi-hop} and Knowledge Base Population (KBP) \citep{lin2020kbpearl}. 

\begin{figure}[!t]
\centering
\includegraphics[width=.48\textwidth]{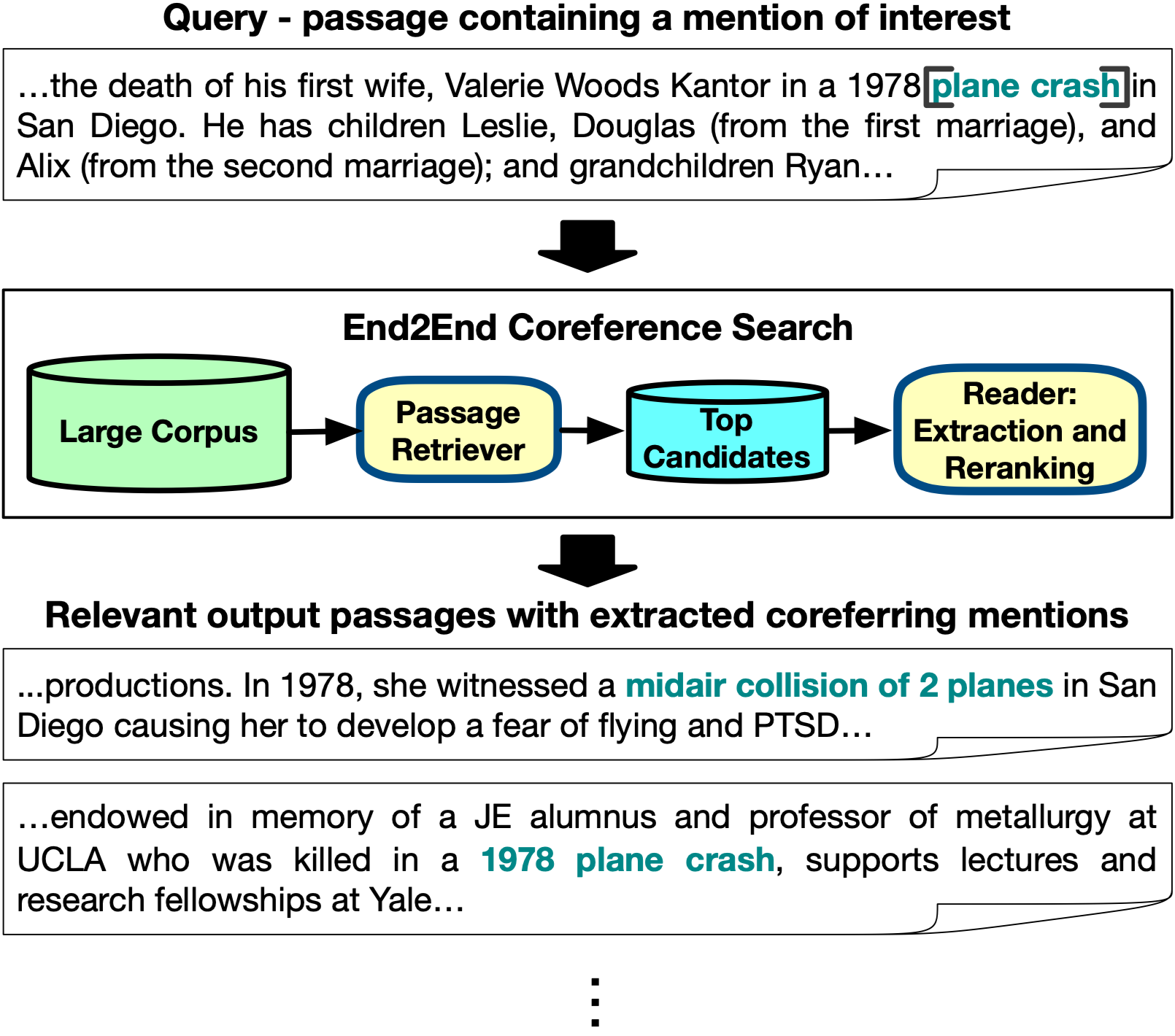}
\caption{Example of Coreference Search. Provided with a query passage containing a mention of interest, a coreference search system retrieves from a large corpus the best candidate passages containing mentions coreferring with the query.}
\label{fig:figure_arch}
\end{figure}

Currently, the CDEC task, as formed in corresponding datasets, is intended at creating models that exhaustively resolve all coreference links in a given dataset. However, an applicable realistic scenario may require to efficiently search and extract coreferring events of only specific events of interest. 
A typical such use-case can be of a user reading a text and encountering an event of interest (for example, the \textit{plane crash} event in Figure~\ref{fig:figure_arch}), and then wishing to further explore and learn about the event from a large document collection.

To address such needs, we propose an appealing, and often more applicable, complementary set up for the task -- \textit{Cross-document Coreference Search} (Figure~\ref{fig:figure_arch}), focusing in this paper on event coreference. Concretely, given a mention in context of an event of interest, considered as a query, the task is to find all coreferring mentions for the query event in a large corpus.

Such coreference resolution search use-case cannot be addressed currently, for two main reasons: (1) Existing CDEC datasets are relatively small for the realistic representation of a search task; (2) Current CDEC models, which are designed at linking all coreference links in a given dataset, are inapplicable in terms of computation at the much larger search space required by realistic coreference resolution search scenarios.



To facilitate research on this setup, we present a large dataset, derived from Wikipedia, by leveraging existing annotations in the Wikipedia Event Coreference dataset (WEC) \cite{eirew-etal-2021-wec}. Our curated dataset resembles in structure to an Open-domain QA (ODQA) dataset \cite{berant-etal-2013-semantic, baudi-etat-2015, joshi-etal-2017-triviaqa, kwiatkowski-etal-2019-natural, rajpurkar-etal-2016-squad}, containing a set of coreference queries and a large passage collection for retrieval.

Observing that the coreference search setup is largely analogous to the setting of Open Domain Question Answering, we adapt the prominent Deep Passage Retrieval (DPR) model to our setting, as an appealing baseline.
Further, motivated to integrate coreference modeling into DPR,
we adapted components inspired by a prominent within-document end-to-end coreference resolution model \cite{lee-etal-2017-end}, which was previously applied also to the CDEC task \cite{cattan2020streamlining}. 
Thus, we developed an integrated model that leverages components from both DPR and the coreference model of \citet{lee-etal-2017-end}. Our novel model yields substantially improved performance on several important evaluation metrics. 

Our dataset\footnote{\url{https://huggingface.co/datasets/Intel/CoreSearch}} and code\footnote{\url{https://github.com/AlonEirew/CoreSearch}} are released for open access.

\section{Background}
In this section, we first describe the Cross Document Event Coreference (CDEC) task, datasets and models (§\ref{sec:cdec-datasets}) and then review the common open-domain QA model architecture (§\ref{sec:qa-models}).

\subsection{Cross-Document Event Coreference Resolution}
\label{sec:cdec-datasets}

ECB+ \cite{cybulska-vossen-2014-using} is the most commonly used dataset for training and testing models for cross-document event coreference resolution. This corpus consists of documents partitioned into 43 clusters, each corresponding to a certain news topic. 
ECB+ is relatively small, where on average only 1.9 sentences per document were selected for annotation, yielding only 722 non-singleton coreference clusters in total (that is, clusters containing more than a single event mention, while singleton clusters correspond to mentions that do not hold a coreference relation with any other mention in the data). 

Since annotating a CDEC dataset is a very challenging task, several annotation methods try to semi-automatically create a CDEC dataset by taking advantage of available resources. The Gun Violence Corpus (GVC) \cite{vossen-etal-2018-dont} leveraged a structured database recording gun violence events for creating an annotation scheme for gun violence related events. In total GVC annotated 7,298 mentions distributed into 1,046 non-singleton clusters.

More recently, \wecEng{} \cite{eirew-etal-2021-wec} and HyperCoref \cite{bugert-gurevych-2021-event} leveraged article hyperlinks pointing to the same concept in order to create an automatic annotation process. 
This annotation scheme helped HyperCoref curate 2.7M event mentions distributed among 0.8M event clusters, extracted from news articles. The smaller \wecEng{} curates 43,672 event mentions distributed among 7,597 non-singleton clusters. Differently then HyperCoref, the \wecEng{} development set (containing 1,250 mentions and 233 clusters) and test set (contains 1,893 mentions and 322 clusters) have gone through a manual validation process (see Table~\ref{tab:wecEng-stats}), ensuring their high quality.


All the above mentioned datasets are targeted for models which exhaustively resolve all coreference links within a given dataset \cite{barhom-etal-2019-revisiting, meged-etal-2020-paraphrasing, cattan2020streamlining, caciularu-etal-2021-cdlm-cross, yu2020paired, held-etal-2021-focus, allaway-etal-2021-sequential, hsu2022contrastive}. This setting resembles the within-document coreference resolution setting, where similarly all links are exhaustively resolved in a given single-document. However, while within-document coreference resolution is contained to a single document, CDCR might relate to an unbounded multi-text search space (e.g., news articles, Wikipedia articles, court and police records and so on). To that end, we aim at a task and dataset for modeling CDEC as a search problem. To facilitate a large corpus for a realistic representation of such a task, while ensuring reliable development and test sets, we adopted the \wecEng{}\footnote{The larger magnitude of HyperCoref makes it a suitable candidate for our \weces{}. However, since HyperCoref is not publicly released, we could not evaluate on it. We leave this part to future work. } as the basis for our dataset creation (§\ref{sec:data_create}). 

\paragraph{Within Document Coreference Resolution}
Recent within-document coreference resolution models \cite{lee-etal-2018-higher, joshi-etal-2019-bert, kantor-globerson-2019-coreference, wu-etal-2020-corefqa}, were inspired by the end-to-end model architecture introduced by \citet{lee-etal-2017-end}. In particular, two distinct components were adopted in those works, which were shown to be effective in detecting mentions and their coreference relations, both in the within-document and cross-document \cite{cattan2020streamlining} settings. In our proposed model, we similarly adopt those two components to better represent coreference relations, in the coreference search settings.

\subsection{Open-Domain Question Answering}
\label{sec:qa-models}



Open-domain question answering (ODQA) \cite{Voorhees99thetrec-8}, is concerned with answering factoid questions based on a large collection of documents. 
Modern open-domain QA systems have been restructured and simplified by combining information retrieval (IR) techniques and neural reading comprehension models \cite{chen-etal-2017-reading}. In those approaches, a retriever component finds documents that might contain an answer from a large collection of documents, followed by a reader component that finds a candidate answer in a given document \cite{lee-etal-2019-latent, yang-etal-2019-end-end, karpukhin-etal-2020-dense}. 

We observe that the Cross-Document Event Coreference Search (CDES) setting resembles the ODQA task. Specifically, given a passage containing a mention of interest, considered as a \emph{query}, CDES is concerned with finding mentions coreferring with the query event in a large document collection. To facilitate research in this task, we created a dataset similar in structure to ODQA datasets \cite{berant-etal-2013-semantic, baudi-etat-2015, joshi-etal-2017-triviaqa, kwiatkowski-etal-2019-natural, rajpurkar-etal-2016-squad}, and established a suitable model resembling in architecture to the recent two-step (retriever/reader) systems, as described in the following sections. 

\begin{table}[!t]
    \centering
    \resizebox{0.4\textwidth}{!}{
    \begin{tabular}{@{}lccc@{}}
        \toprule
        & \makecell{Mentions} & \makecell{None-Singleton \\ Clusters} \\
        \midrule
        \wecEng{} (train) & 40,529 & 7,042 \\
        \wecEng{} (dev) & 1,250 & 216 \\
        \wecEng{} (test) & 1,893 & 306 \\
        \bottomrule
    \end{tabular}}
    \caption{WEC-Eng Dataset Statistics. \textbf{Mentions}: The total number of event mentions within the corresponding section. \textbf{Non-Singleton Clusters}: Number of event clusters containing more than a single event mention.}
    \label{tab:wecEng-stats}
\end{table}

\begin{table}
    \centering
    \resizebox{0.48\textwidth}{!}{
    \begin{tabular}{@{}lcccc@{}}
        \toprule
        & \makecell{Train} & \makecell{Dev} & \makecell{Test} & \makecell{Total} \\
        \midrule
        \wecEng{} Validated Data & & & & \\
        \quad \# Clusters & 237 & 49 & 236 & 522 \\
        \quad \# Passages (with Mentions) & 1,503 & 341 & 1,266 & 3,110 \\
        \midrule
        \# Added Destructor Passages & 922,736 & 923,376 & 923,746 & 2,769,858 \\
        \midrule
        \# Total Passages & 924,239 & 923,717 & 925,012 & 2,772,968 \\
        \bottomrule
    \end{tabular}}
    \caption{\weces{} dataset statistics.}
    \label{tab:weces-stats}
\end{table}

\section{The \weces{} Dataset}
\label{sec:data_create}
We formulated the \textit{Cross-Document Event Coreference Search} task following a similar approach to open-domain question answering (illustrated in Figure~\ref{fig:figure_arch}). Specifically, given a query containing a marked target event mention, along with a passage collection, the goal is to retrieve all the passages from the passage collection that contain an event mention coreferring with the query event, and extract the coreferring mention span of each retrieved passage. 

To facilitate research on this task, we present a large dataset, derived from Wikipedia, termed \textit{\weces{}}. In this section we describe the \weces{} dataset structure (§\ref{sec:composit}), following by describing the structure of a single query instance (§\ref{sec:instance}).

\subsection{Dataset Structure}
\label{sec:composit}
The \weces{} dataset consists of two separate passage collections: (1) a collection of passages containing manually annotated coreferring  event mention, and (2) a collection of destructor passages. 

\paragraph{Annotated Data}
The \weces{} passage collection which contains manually annotated event mentions was created by importing the validated portion of the \wecEng{} \cite{eirew-etal-2021-wec} dataset (§\ref{sec:cdec-datasets} and Table~\ref{tab:wecEng-stats}). 

Specifically, we merged the \wecEng{} validated test and development set coreference clusters into a single collection of 522 none-singleton clusters (``Non-Singleton Culsters'' in Table~\ref{tab:wecEng-stats} and ``Clusters'' in Table~\ref{tab:weces-stats}).
We then split the clusters between \weces{} train, development and test sets. Each cluster contains passages that form our annotated passage collection. 

Those passages will serve the roles of queries and of positive retrieved coreferring passages.

\paragraph{Destructor Passages}
In order to collect a large collection of passages for challenging and realistic retrieval, we generate negative passages (i.e., destructing passages) using two resources: (1) The entire \wecEng{} train set, which is not manually validated, though quite reliable; (2) By extracting the first paragraph of any Wikipedia article not containing a hyperlink to any of the \weces{} annotated passages, and hence are unlikely to corefer with any of them (Table~\ref{tab:weces-stats}).

\paragraph{Cluster Types}
We observe that our annotated data is characterized by two prominent types of coreference clusters: \textit{Type-1} - clusters containing only passages with event mention spans that include the event time or location (e.g., ``2006 Dahab bombings'', ``2013's BET Awards''), and \textit{Type-2} - clusters that are comprised partly of passages as in Type-1, as well as passages containing mention spans without any event identifying participants (e.g., ``the deadliest earthquake on record'', ``BET Awards'', ``plane crash''). Naturally, Type-2 clusters will create queries/passage examples with a higher degree of difficulty. Identifying coreference for Type-2 clusters is indeed challenging in our dataset, because \wecEng{} includes a multitude of event mentions which are similar lexically but do not corefer (e.g., different earthquakes) \cite{eirew-etal-2021-wec}, requiring a model to identify event coreference using the arguments in the surrounding context.

To measure the distribution of cluster types within \weces{}, we randomly sampled 20 clusters and found 90\% are of type-2, demonstrating the challenging nature of the \weces{} data. Table~\ref{tab:clusters} illustrates examples of queries extracted randomly from five type-2 clusters.

\begin{table}[!t]
    \centering
    \resizebox{0.48\textwidth}{!}{
    \begin{tabular}{@{}p{5cm}p{5cm}@{}}
        \toprule
        \makecell{Query} & \makecell{Unique Positive \\ Passage Mention Answers} \\
        \midrule
        ...On 14 April 2010, an \textbf{\textcolor{greenrgb}{earthquake}} struck the prefecture, registering a magnitude of 6.9... & `Yushu earthquake', `2010 Yushu earthquake', `earthquake in Qinghai', `earthquake in 2010', `Qinghai earthquake' \\
        \midrule
        ...2012–13 season 10–21, 6–12 in MAAC play to finish in eighth place. They lost in the first round of the \textbf{\textcolor{greenrgb}{MAAC Tournament}} to... & `\textcolor{blue}{MAAC Tournament}', `2013 MAAC Tournament' \\
        \midrule
        ...Salo and the band The Ark won Melodifestivalen 2007 and went on to represent Sweden in the \textbf{\textcolor{greenrgb}{Eurovision Song Contest}} 2007 with the song... & `52nd Eurovision Song Contest 2007', `\textcolor{blue}{previous contest}', `2007 edition of the Contest', `2007 contest', `\textcolor{blue}{that year's contest}', `Eurovision 2007' \\
        \midrule
        ...finished the season 18–15, 10–8 in Pac-10 play. They lost to USC in the quarterfinals \textbf{\textcolor{greenrgb}{Pac-10 tournament}}... & `2011 Pacific-10 Conference Men's Basketball Tournament', `2011 Pac-10 Tournament', `2011 Pac-10 tournament' \\
        \midrule
        ...The film was planned to premiere at the 65th annual \textbf{\textcolor{greenrgb}{Cannes International Film Festival}} in May 2012, but in late 2011  & `\textcolor{blue}{Cannes Short Film Corner}', `2012 Cannes Film Festival', `\textcolor{blue}{Cannes Film Festival}', `65th Annual Cannes Film Festival' \\
        \bottomrule
    \end{tabular}}
    \caption{Sample of five queries containing a mention (highlighted in \textbf{\textcolor{greenrgb}{green}}) without event participants, and the corresponding cluster mentions (\textcolor{blue}{blue} highlights passages mentions without event participants), illustrating challenging query examples in \weces{} dataset.}
    \label{tab:clusters}
\end{table}

\subsection{CoreSearch Instance Structure}
\label{sec:instance}
An instance in the \weces{} dataset is comprised of: (1) a query passages pulled from the annotated passage collection; (2) The collection of all other passages, which are considered as the passage collection for retrieval. Passages in the passage collection which belong to the same cluster as the pulled query are considered positive passages, while all the rest as negative passages.

\paragraph{Potential Language Adaptation}
The \weces{} dataset is built on top of the English version of WEC (\wecEng{}). Consequently, since WEC is adoptable to other languages with relatively low effort \cite{eirew-etal-2021-wec}, and the process for deriving \weces{} from it is simple and fully automatic, the \weces{} dataset may be adopted to other languages as well with very similar effort (as for WEC).

\section{Coreference-search Models} 
In this section, we aim to devise an effective baseline for our event coreference search task to be trained
on our dataset. Following the observation that coreference search formulation resembles the open-domain QA (ODQA) (§\ref{sec:qa-models}), we propose an end-to-end neural architecture, comprised of a \emph{retriever} and a \emph{reader} models. Given a query passage, the retriever selects the top-$k$ most relevant passage candidates out of the entire passage corpus (§\ref{sec:retriever-model}). Then, the reader is responsible for re-ranking the retrieved passages and extracting the coreferring event span, by using a reading comprehension module (§\ref{sec:reader-model}). 

\subsection{The Retriever Model}
\label{sec:retriever-model}
Given a query passage containing an event mention of choice, the goal of the \emph{retriever} is to select the top-$k$ relevant passage candidates out of a large collection of passages. To that end, we build upon the foundations of the Dense Passage Retriever model \cite{karpukhin-etal-2020-dense} and employ a similar retriever.

Similarly to DPR, we propose to encode the query passage $q_i=[\texttt{CLS}, q_i^1, \ldots, q_i^{n_i}]$ and a candidate passage $p_j=[\texttt{CLS}, p_j^1, \ldots, p_j^{n_j}]$ using two distinct neural encoders, $E_Q(\cdot)$ and $E_P(\cdot)$,\footnote{As in DPR, after training these encoders, we use $E_P(\cdot)$ to build an index for all the passages in the corpus prior to applying the test-time retrieval.} for mapping their tokens into $d$-dimensional dense vectors, $[\textbf{q}_i^\texttt{CLS}, \textbf{q}_i^1, \ldots, \textbf{q}_i^{n_i}]$ and $[\textbf{p}_j^\texttt{CLS}, \textbf{p}_j^1, \ldots, \textbf{p}_j^{n_j}]$ for $q_i$ and $p_j$, respectively. Here, both $\textbf{q}_i^\texttt{CLS}$ and $\textbf{p}_j^\texttt{CLS}$ denote the last hidden layer contextualized $\texttt{[CLS]}$ token representations of $q_i$ and $p_j$ respectively, which are then fed to a dot-product similarity scoring function, which determines candidate passage ranking:

\begin{equation}
    \text{sim}(q_i, p_j) =\textbf{q}_i^\texttt{CLS}\cdot \textbf{p}_j^\texttt{CLS}
    \label{eq:sim}
\end{equation}


\paragraph{Event Mention Marking} In order to accommodate our setup of mention-directed search and to better signal the model to be aware of the query event mention, we edit the query by marking the span of the mention within the query passage by using boundary tokens. Given the query event mention span $m_i=[q_i^k, q_i^{k+1},\ldots,q_i^{k+l-1}]$, we append the boundary tokens to obtain the final edited query ($m_i$ denotes the sequence of the mention's tokens):
$$q_i=[\texttt{CLS},\ldots,q_i^{k-1},\langle\texttt{S}\rangle,m_i,\langle\setminus\texttt{S}\rangle,q_i^{k+l},\ldots q_i^{n_i}].$$

\paragraph{Improved Span Representation} For implementing the text encoders $E_Q(\cdot)$ and $E_P(\cdot)$, we employed the SpanBERT\footnote{DPR originally used BERT \cite{devlin-etal-2019-bert} as their query and passage encoders.} \cite{joshi-etal-2020-spanbert} model as our query and passage encoders. SpanBERT is an appealing encoder, as it was pre-trained for better span representations, rather then the individual tokens, and was also shown to be more effective for coreference resolution tasks \cite{joshi-etal-2020-spanbert,wu-etal-2020-corefqa}. 

During our preliminary experiments, we observed that both the additional event mention marking as well as replacing BERT with SpanBERT contributed significantly to the performance over our dataset.

\paragraph{Positive and Negative Training Examples}
We construct our positive and negative examples by iterating sequentially through every training set event coreference cluster $C_j=[m_1, m_2,\ldots, m_{|C_j|}]$, where $m_i$ denotes an event mention surrounded with its context (the entire passage). Given each event mention $m_i$ acting as a query $q_i$, we construct one positive coreference example for each of the remaining $|C_j|-1$ coreferring event mentions in the cluster. Then, for each such positive example, we first construct one ``challenging" negative example by selecting randomly one of the top-20 passages returned by the BM25 retrieval model for the corresponding query. In addition, for each query in a training batch, we create additional (``easier") in-batch negative examples by taking the ``challenging" passages of all other queries in the current batch, similarly to \citet{karpukhin-etal-2020-dense}.



\paragraph{Objective} Let ${\mathcal{D}={\langle q_i, p^+_i, p^-_{i,1},\ldots, p^-_{i,n} \rangle}^m_{i=1}}$ be the \weces{} training set. Similarly to \citet{karpukhin-etal-2020-dense}, the goal is to optimize the negative log likelihood loss of the positive passage, which is based on the contrastive loss:
\begin{equation}
    \begin{gathered}[b]
        \mathcal{L}\left(q_i, p^+_i, p^-_{i,1},\ldots, p^-_{i,n}\right)\\ =-\log\frac{e^{\text{sim}(q_i,p_i^+)}}{e^{\text{sim}(q_i,p_i^+)} + \sum_{j=1}^{n}e^{\text{sim}(q_i,p_i^-)}}.
    \end{gathered}
\end{equation}




\subsection{The Reader Model}
\label{sec:reader-model}
Given a mention surrounded by its context as the query, and its top-$k$ retrieved passages, the reader model is tasked to (1) re-rank the retrieved passages according to a passage selection score and (2) extract the candidate mention span from each passage. We implemented two flavours of readers, a DPR baseline (§\ref{sec:model-dpr-read-baseline}), and a DPR reader enhanced with event coreference scores (§\ref{sec:model-integrated-read}). 


\subsubsection{DPR Reader Baseline} 
\label{sec:model-dpr-read-baseline}
We implemented a DPR-based passage selection model that acts as re-ranker through cross-encoding the query and the passage. Specifically, we append a query $q_i$ (including the event mention marker tokens, see §\ref{sec:retriever-model}) and a passage $p_j$, and feed the concatenated input sequence to the RoBERTa text encoder $E_R(\cdot)$ \cite{liu-2019-RoBERTaAR}. Similarly to \citet{karpukhin-etal-2020-dense}, we then use the output (last hidden layer) token representations to predict three probability distributions. We compute the span score of the $s^\text{th}$ to $t^\text{th}$ tokens from the $j^\text{th}$ passage as $P_{\text{start},j}(s) \times P_{\text{end},j}(t)$, and a passage selection score of the $j^\text{th}$ passage as $P_{\text{select}}(j)$:

\begin{flalign} 
    P_{\text{start},j}(s) &= \text{softmax}(\mathbf{P}_j\mathbf{w}_{\text{start}})_s & \label{eq:start} \\
    P_{\text{end},j}(t) &= \text{softmax}(\mathbf{P}_j\mathbf{w}_{\text{end}})_t & \label{eq:end} \\
    P_{\text{select}}(j) &= \text{softmax}(\mathbf{\hat{P}}^T\mathbf{w}_{\text{select}})_j, &
    \label{eq:select}
\end{flalign}


where $[\cdot]$ denotes column concatenation, \begin{small}${\mathbf{P}_j = [\textbf{p}_j^{\texttt{CLS}},\textbf{p}_j^{1}, \ldots, \textbf{p}_j^{n_j}]}$\end{small}, \begin{small}${\mathbf{\hat{P}} = [\textbf{p}^{\texttt{CLS}}_1, \ldots, \textbf{p}^{\texttt{CLS}}_k]}$\end{small}, $k$ is the number of the retrieved passages, and ${\mathbf{w}_{\text{start}}, \mathbf{w}_{\text{end}}, \mathbf{w}_{\text{select}}}$ are learned vectors.


\subsubsection{Integrating the Coreference Signal} 
\label{sec:model-integrated-read}
While the above DPR-based reader yields appealing performance (§\ref{sec:e2e-results}), we conjecture that the passage selection (Eq. \ref{eq:select}), which is based on the passages' $\texttt{[CLS]}$ token representations, is sub-optimal for coreference resolution. These representations learn high-quality sentence- or document-level features \cite{devlin-etal-2019-bert}, however in our setting, more fine-grained features are required in order to capture information for better modeling coreference relations between mention spans. Motivated by this hypothesis, we replaced the passage selection component (Eq. \ref{eq:select}) with a method adapted from recent neural within-document coreference models \cite{lee-etal-2017-end, lee-etal-2018-higher, joshi-etal-2019-bert, kantor-globerson-2019-coreference, wu-etal-2020-corefqa}. 

Specifically, we aim to model the probability of passage $j$ to be selected by the likelihood it contains an event mention $m_j$ that corefers to the query's event mention $m_i$:
\begin{flalign}
    P_{select}\left(j\right) &= \frac{e^{s\left(m_j,m_i\right)}}{\sum_{j=1}^{k}e^{s\left(m_j,m_i\right)}} \\
    s\left(m_j,m_i\right) &= s_m\left(m_j\right) + s_a\left(m_j,m_i\right) &\\
    s_m\left(m_j\right) &= w_m \cdot \texttt{FFNN}_m\left(g_j\right) &\\
    s_a\left(m_j,m_i\right) &= w_a\cdot \texttt{FFNN}_a\left(\left[g_i,g_j,g_i\circ g_j\right]\right)
\end{flalign}


Where $s_m(m_j)$ is the mention scorer, $s_a(m_j,m_j)$ is the antecedent scorer that computes coreference likelihood for the pair of mentions, $\circ$ represents the element-wise product of $g_j$ and $g_i$, $g_x=[m_{x,s}, m_{x,t}]$ is the concatenated vector of the first and last token representations of the mention in the passage $x\in\{i,j\}$, and $s(m_i, m_j)$ is the final pairwise score. \texttt{FFNN} represents a feed forward neural network with a single hidden layer. Note that standard coreference resolution methods compute also $s_m(m_i)$, however since in our setup the query mention is constant, it can be omitted.   

During training, we extract the gold start/end embeddings of the candidate passage, while at inference time, we use the scores computed by Eq.~\ref{eq:start} and Eq.~\ref{eq:end} (see §\ref{sec:model-dpr-read-baseline}) in order to extract the most probable plausible mention spans. Invalid spans, who's end precedes their start position point or are longer than a threshold $L$, are filtered. For the query event mention, we use the same mention marking strategy used for the query encoder (§\ref{sec:retriever-model}). We further show in §\ref{sec:e2e-results} that this marking improves the performance of the reader.






\section{Experiments and Results}
\label{sec:exp}

\subsection{Implementation Details}
\paragraph{Retriever}
We train the two separate encoders using a maximum query size of 64 tokens for the query encoder. In order to cope with memory constrains, we limit the maximum passage size given to the passage encoder to 180 tokens. Batch size is set to 64. We train our model using four 12GB Nvidia Titan-Xp GPUs.\footnote{We leveraged and modified the implementations in the Haystack framework of the DPR and BM25 models: \url{https://github.com/deepset-ai/haystack}}

\paragraph{Reader}
We train the single cross-encoder using a maximum sequence size of 256 tokens, in order to cope with memory constraints. We use up to 64 tokens from the surrounding query mention context (which in many cases take less then 64 tokens) for query representation, and concatenate the passage context using the remaining available sequence. In case the passage context length exceeds available sequence size for passage representation, we segment the passage using overlapping strides of 128 tokens, creating additional passage instances with the same query. The batch size is set to 24, and both $\texttt{FFNN}_m$, $\texttt{FFNN}_a$ use a single hidden layer set to 128. We train the models using two 12GB Nvidia Titan-Xp GPUs.

\paragraph{Hyperparameters}
All models parameters are updated by the AdamW \cite{loshchilov201-decoupled-wd} optimizer, with a learning rate set to $10^{-5}$ and a wight-decay rate of 0.01. We also apply a linear scheduling with warm-up (for 10\% of optimization steps) and dropout rate of 0.1. We train all models for 5 epochs and consider the best performing ones over the development set. At inference, we set the retriever top-$k$ parameter to 500. 



\subsection{Evaluation Measures}
\label{sec:eval}
In all our experiments, we followed the common evaluation practices used for evaluating Information Retrieval (IR) models \cite{Khattab2020ColBERTEA, xiong2021approximate, hofstatter-etal-2021, thakur2021beir}. Accordingly, we used the following metrics:

\paragraph{Mean Reciprocal Rank (MRR@$k$)} 
Following common evaluation practices, we set $k$ to 10, expecting that the topmost correct result should appear amongst the top 10 results (that is, no credit is given if the topmost correct result is ranked lower than 10).

\paragraph{Recall (R@$k$)} 
We report recall at ${k\in\{10, 50\}}$ for the end-to-end model evaluation, assessing recall in two prototypical cases where the user might choose to look at rather few or rather many results. For the retriever model we report recall at ${k\in\{10, 100, 500\}}$, illustrating the motivation for the $k=500$ cutoff point that we chose (beyond which there were no substantial recall gains).

\paragraph{mean Average Precision (mAP@$k$)} 
The mAP metric assesses the ranking quality of \textbf{all} correct results within the top-k ones, measured for $k\in\{10, 50\}$, as measured for recall.\footnote{We use mAP rather than Normalized Discounted Cumulative Gain (NDCG), because the latter requires a scaled gold relevancy score for each query result.
mAP applies a similar ranking evaluation criterion, but is suitable for binary relevancy scores, which is the case in our coreference setting.}



\paragraph{Reader Evaluation} We use the above metrics with the additional question answering (QA) measurements of Exact Match (EM) and token level F1 score\footnote{Using the official SQuAD evaluation script.} with the reference answer after minor normalization as in \cite{chen-etal-2017-reading, lee-etal-2019-latent, karpukhin-etal-2020-dense}.\footnote{We note that QA measurements only take into consideration lexical matches. However, equal lexical representation does not necessarily imply a coreference relation (for example, the mention \textit{plane crash} can appear twice in the same passage, each time referring to a different plane, thus denoting different events). To that end, we add a necessary constraint limiting relevant answers only to those where the answer index within context intersects with the gold mention span.}

\begin{table}
    \centering
    \resizebox{0.48\textwidth}{!}{
    \begin{tabular}{@{}lcccccc@{}}
        \toprule
        \makecell{\textbf{Model}} & \makecell{\textbf{MRR@10}} & \makecell{\textbf{mAP@10}} & \makecell{\textbf{mAP@50}} & \makecell{\textbf{R@10}} & \makecell{\textbf{R@100}} & \makecell{\textbf{R@500}} \\
        \midrule
        \midrule
        \multicolumn{7}{c}{Development} \\
        \midrule
        BM25 & 57.32 & 25.92 & 31.05 & 27.63 & 56.09 & 74.75 \\
        $\mbox{Retriever-B}^-$ & 23.56 & 4.91 & 7.71 & 9.53 & 42.66 & 65.09 \\
        $\mbox{Retriever-S}^-$ & 75.4 & 37.6 & 44.09 & 40.06 & 71.65 & 86.21 \\
        $\mbox{Retriever-S}^+$ & \textbf{80.92} & \textbf{40.43} & \textbf{47.5} & \textbf{41.59} & \textbf{74.03} & \textbf{87.53} \\
        \midrule
        \multicolumn{7}{c}{Test} \\
        \midrule
        BM25 & 62.45 & 29.75 & 34.02 & 27.82 & 57.9 & 74.75 \\
        $\mbox{Retriever-S}^+$ & \textbf{69.1} & \textbf{35.73} & \textbf{43.24} & \textbf{37.44} & \textbf{75.31} & \textbf{87.12} \\
        \bottomrule
    \end{tabular}}
    \caption{Retriever results on \weces{} development and test sets. \textbf{BM25}: BM25 score; \textbf{$\mbox{Retriever-B}^-$}: DPR retriever using BERT, without mention boundary tokens; \textbf{$\mbox{Retriever-S}^-$}: DPR retriever using SpanBERT, without boundary tokens; \textbf{$\mbox{Retriever-S}^+$}: Our complete retriever, with boundary tokens}
    \label{tab:ret-results}
\end{table}

\begin{table}
    \centering
    \resizebox{0.48\textwidth}{!}{
    \begin{tabular}{@{}lccccccc@{}}
        \toprule
        \makecell{\textbf{Model}} & \makecell{\textbf{MRR@10}} & \makecell{\textbf{mAP@10}} & \makecell{\textbf{mAP@50}} & \makecell{\textbf{R@10}} & \makecell{\textbf{R@50}} & \makecell{\textbf{EM}} & \makecell{\textbf{F1}} \\
        \midrule
        \midrule
        \multicolumn{8}{c}{Development} \\
        \midrule
        $\mbox{E2E-DPR}^-$ & 89.81 & 58.19 & 68.35 & 55.16 & \textbf{84.47} & 74.99 & 82.97 \\
        $\mbox{E2E-DPR}^+$ & 92.48 & 58.23 & 65.14 & 53.34 & 73.81 & 79.59 & \textbf{88.91} \\
        E2E-Integrated & \textbf{94.05} & \textbf{61.82} & \textbf{70.81} & \textbf{56.53} & 82.19 & \textbf{83.31} & 88.78 \\
        \midrule
        \multicolumn{8}{c}{Test} \\
        \midrule
        $\mbox{E2E-DPR}^-$ & 87.93 & 60 & 70.51 & 52.3 & \textbf{86.62} & 65.35 & 77.77 \\
        $\mbox{E2E-DPR}^+$ & 88.18 & 58.26 & 66.37 & 49.24 & 76.66 & 69.87 & 82.92 \\
        E2E-Integrated & \textbf{90.06} & \textbf{63.26} & \textbf{72.91} & \textbf{53.5} & 84.35 & \textbf{71.44} & \textbf{84.16} \\
        \bottomrule
    \end{tabular}}
    \caption{End-to-end results on \weces{} development and test sets. \textbf{E2E-DPR}: the end-to-end DPR baseline results, where '$^-$' indicates the model was trained without mention boundary tokens, and '$^+$' with them. \textbf{E2E-Integrated}: Our end-to-end integrated model.}
    \label{tab:read-results}
\end{table}

\begin{table*}
    \centering
    \resizebox{1\textwidth}{!}{
    \begin{tabular}{@{}c|c|c|c@{}}
    \toprule
    \makecell{\textbf{No}} & \makecell{\textbf{Query Context}} & \makecell{\textbf{Top-2 Results}} & \makecell{\textbf{Relevancy}}  \\
    \toprule
        1 & 
        \begin{tabular}{@{}p{10cm}@{}} 
            ...Walid al-Maqdisi, a Salafi leader of an al-Qaeda-affiliated terrorist group, responsible for \textbf{\textcolor{blue}{three bombings}} in Dahab in 2006, and which is believed to have close ties with terror cells operating in the Sinai Peninsula...
        \end{tabular}
        & 
        \begin{tabular}{@{}p{10cm}@{}}
            ...to replace it with other measures, such as specific anti-terrorism legislation. The extension was justified by the \textbf{\textcolor{greenrgb}{Dahab bombings}} in April of that year...
            \\\midrule
            ...damaging the industry so that the government would pay more attention to their situation. (See 2004 Sinai bombings, 2005 Sharm El Sheikh bombings and \textbf{\textcolor{greenrgb}{2006 Dahab bombings}})...
        \end{tabular}
        & 
        \makecell{\textbf{\textcolor{greenrgb}{\cmark}} \\\\\\ \textbf{\textcolor{greenrgb}{\cmark}}}
        \\
        \midrule
        2 & 
        \begin{tabular}{@{}p{10cm}@{}} 
            ...On 14 April 2010, an \textbf{\textcolor{blue}{earthquake}} struck the prefecture, registering a magnitude of 6.9 (USGS, EMSC) or 7.1 (Xinhua). It originated in the Yushu Tibetan Autonomous Prefecture, at local time...
        \end{tabular}
        & 
        \begin{tabular}{@{}p{10cm}@{}} 
            ...The airport played an important role in the delivery of rescue personnel and relief supplies to the area affected by the \textbf{\textcolor{greenrgb}{2010 Yushu earthquake}}... 
            \\\midrule
            ...China-Congo Friendship Primary School, a school mostly for Tibetan orphans in Chindu County, Qinghai, after the \textbf{\textcolor{greenrgb}{2010 Yushu earthquake}} destroyed the old school...
        \end{tabular}
        & 
        \makecell{\textbf{\textcolor{greenrgb}{\cmark}} \\\\\\ \textbf{\textcolor{greenrgb}{\cmark}}}
        \\
        \midrule
        3 & 
        \begin{tabular}{@{}p{10cm}@{}} 
            ...He made his AFL debut in the 2010 season and was rewarded with an \textbf{\textcolor{blue}{AFL Rising Star}} nomination. He spent six seasons with Essendon, which peaked with a fifth-place finish in the best and fairest, and after 114 games with the club, he was traded to the Melbourne Football Club during the 2015 trade period...
        \end{tabular}
        & 
        \begin{tabular}{@{}p{10cm}@{}} 
            ...Aaron Joseph was nominated for the \textbf{\textcolor{purple}{2009 AFL Rising Star}} award for his performance in Carlton's Round 12 win against. Joseph did not poll votes in the final count...
            \\\midrule
            ...Davis made his AFL debut for Adelaide in Round 4, 2010 against Carlton at AAMI Stadium; he had 16 possessions and seven marks. Davis was nominated for the \textbf{\textcolor{greenrgb}{2010 Rising Star}} in round 16...
        \end{tabular}
        & 
        \makecell{\textbf{\textcolor{purple}{\xmark}} \\\\\\ \textbf{\textcolor{greenrgb}{\cmark}}}
        \\
        \midrule
        4 & 
        \begin{tabular}{@{}p{10cm}@{}} 
            ...In the Tang Dynasty, 10 emperors were buried in Weinan after their death. On the morning of 23 January 1556, \textbf{\textcolor{blue}{the deadliest earthquake on record}} with its epicenter in Huaxian killed approximately 830,000 people...
        \end{tabular}
        & 
        \begin{tabular}{@{}p{10cm}@{}} 
            ...including the \textbf{\textcolor{greenrgb}{1556 Shaanxi earthquake}} that reportedly killed more than 830,000 people, listed as the deadliest earthquakes of all times and the third deadliest natural disaster...
            \\\midrule
            ...In 1556, during the rule of the Jiajing Emperor, the \textbf{\textcolor{greenrgb}{Shaanxi earthquake}} killed about 830,000 people, the deadliest earthquake of all time...
        \end{tabular}
        & 
        \makecell{\textbf{\textcolor{greenrgb}{\cmark}} \\\\\\ \textbf{\textcolor{greenrgb}{\cmark}}}
        \\
    \bottomrule
    \end{tabular}}
    \caption{The top-2 query results given by the E2E-Integrated model on a random sample of \textit{Type-2} cluster queries (§\ref{sec:composit}). \textbf{\textcolor{blue}{Blue}} signifies the mention span in the query, \textbf{\textcolor{greenrgb}{green}} signifies a correct mention detection, and \textbf{\textcolor{purple}{purple}} signifies a wrong mention detection. The relevancy indicator column signifies whether the retrieved passage itself is relevant or not.}
    \label{tab:query-examples-sml}
\end{table*}

\subsection{Results}
\paragraph{Retriever}
Table~\ref{tab:ret-results} summarizes the retriever performance results over the \weces{} test set. Our retriever model surpasses the BM25 method (see further details in Appendix~\ref{apd:sparse-retriever}) by a large margin on every metric (Table~\ref{tab:ret-results}, BM25 versus $\mbox{Retriever-S}^+$). It should be noted that BM25 is considered a strong information retrieval model \cite{robertson-etal-2009-prob}, also compared to recent neural-based retrievers \cite{Khattab2020ColBERTEA, izacard-etal-2021-information, chen-et-al-2021-salient, piktus2021web}. 
We observed this phenomenon during our experiments, as the underlying DPR retriever (i.e., BERT without boundary tokens), yielded poor results on our settings, surpassed by the BM25 model on all measurements by a significant gap (Table~\ref{tab:ret-results}, BM25 versus Retriever-B). 

\paragraph{End-to-end}
\label{sec:e2e-results}
Table~\ref{tab:read-results} presents our end-to-end system  results applied over the \weces{} test set. We found that both of the reader models (\textit{E2E-DPR} and \textit{E2E-Integrated}) present appealing performance given different measurement aspects we now describe. 

We conclude from the recall results (R@10 and R@50) that the $\mbox{E2E-DPR}^-$ model is an effective re-ranking model, ranking almost all relevant passages extracted by the retriever within the top 50 results (86.62\% out of maximum of 87.12\% ranked by the $\mbox{Retriever-S}^+$ model at top-500). The EM and F1 results indicate that the E2E-Integrated model gains better mention extraction capabilities compared to both E2E-DPR models (by 1.5\% EM and 1.2\% F1 compared to $\mbox{E2E-DPR}^+$). 

Finally, the MRR and mAP results indicate that the E2E-Integrated model overall performs better then both E2E-DPR models at ranking relevant passages at higher ranks (indicated by MRR@10, mAP@10 and mAP@50 in Table~\ref{tab:read-results}). In particular, we find that the MRR@10 results are especially appealing (90.06\%), showing the model predominately ranks a relevant passage at the first or second position. 

Finally, Table~\ref{tab:query-examples-sml} illustrates a sample of the E2E-Integrated top-2 model results, on a sample of queries containing mention spans not including event arguments, randomly sampled from five \textit{Type-2} \weces{} clusters (§\ref{sec:composit}). The table Illustrates the model effectiveness in returning relevant passages and the coreferring mention within them.

\paragraph{False Negative Passages}
We observed that on rare occasions the model returns a relevant passage (and a coreferring mention) marked as negative in the dataset. We sampled 15 queries and manually validated their top-10 answers. We found that from 58 negative results, only 1 was a false negative, indicating that indeed this phenomenon is rather rare and insignificant. Such false negatives can originate either from the \wecEng{} training set (§\ref{sec:cdec-datasets}), or from our destructing passage generation (§\ref{sec:data_create}). Notice that, such false negatives can only have a deflating effect on results.



\subsection{Ablation Study}
\label{sec:ablation}
To understand further how different model changes affect the results, we conduct several experiments and discuss our findings below. Table~\ref{tab:ret-results} presents the retriever model results and Table~\ref{tab:read-results} presents the reader model results on the development set, for some ablations.

\paragraph{Mention Span Boundaries}


In both our \textit{retriever} and \textit{reader} experiments, we found that adding the span boundary tokens around the query mention, provides a strong signal to the model. In our retriever experiments, while most of the gain to performance was originated by replacing the BERT model with SpanBERT (Retriever-B and $\mbox{Retriever-S}^-$ in Table~\ref{tab:ret-results}), applying boundary tokens significantly improved performance further all across the board ($\mbox{Retriever-S}^+$ in Table~\ref{tab:ret-results}). 

However, in our \textit{end-to-end} model experiments, we observed that applying boundary tokens will help the model mostly to improve at span detection, while less so at re-ranking ($\mbox{E2E-DPR}^-$ and $\mbox{E2E-DPR}^+$ in Table~\ref{tab:read-results}).


\paragraph{Modeling Coreference with QA}

Our main motivation for replacing the DPR reader passage selection method (Eq.~\ref{eq:select}), with a coreference scoring one, was to create a better passage selection mechanism for re-ranking. Indeed, this modeling prove efficient both at re-ranking, as well as at mention detection, as indicated by the E2E-Integrated model results in Table~\ref{tab:read-results}.

\subsection{Qualitative Error Analysis}
To analyze prominent error types made by our E2E-Integrated model we sampled 20 query results that were incorrectly ranked at the first position 
(Table~\ref{tab:err-analysis} in Appendix~\ref{apd:error} presents a few of these examples). From those 20 results, 18 were indeed identified as incorrect while 2 results were actually correct, that is, including a mention that does corefer with the query event but was missed in the annotation (a false negative).

We observed two main errors types. The first type involved event argument inconsistencies, identified in 10 out of the 18 erroneous results. In these cases, the model identified an event of the same type as the query event, 
but with non-matching arguments (see examples 3, 4, 5 and 6 in Table~\ref{tab:err-analysis}).
This type of error suggests that there is room for improving the model capability in within- and cross-document argument matching. 
Some illustrating examples in Table~\ref{tab:err-analysis} for such argument mismatches include ``few days later'',  ``also that year'', ``the town'' (examples 3, 4 and 5, respectively).

The second type of error, identified in 8 out of the 18 erroneous results, corresponded to cases where the two contexts of the query and result passages did not provide sufficient information for determining coreference  (see examples 1 and 2 in Table~\ref{tab:err-analysis}). 
Manually analyzing these 8 cases, we found that in 3 of them the coreference relation could be excluded by examining other event mentions in the coreference cluster to which the query belongs. In 7 cases, it was possible to exclude coreference by consulting external knowledge, specifically Wikipedia, to obtain more information either about the event itself or its arguments.
Example 1 in the table illustrates a case where Wikipedia could provide conflicting information about the event location (the city of the \textit{Maxim restaurant} vs. the city of the query event). Example 2 illustrates a case where Wikipedia provided conflicting information about the event time (the time of the first Republican convention in the query vs. the time of the convention discussed in the result).
This error type suggests the potential for incorporating external knowledge in cross-document event coreference models. Further, models may benefit from  considering globally the information across an entire coreference cluster, as was previously proposed in some works \cite{raghunathan-etal-2010-multi}.

\section{Conclusions}
We introduced \textit{Cross-document Coreference Search}, a challenging task for  accurate semantic search for events. 
To support research on this task, we created the Wikipedia-based \textit{\weces{}} dataset, comprised of training, validation, and test set queries, along with a large collection of about 1M passages to retrieve from in each set. 
Furthermore, our methodology for semi-automatically converting a cross-document event coreference dataset to a coreference search dataset can be applied to other such datasets, for example HyperCoref \cite{bugert-2021} which represents the news domain. Finally, we provide several effective baseline models and encourage future research on this promising and practically applicable task, hoping that it will lead to a broad set of novel applications and use-cases. 


\section{Limitations}


In this work, we construct the \weces{} dataset, which relies on the existing Wikipedia Event Coreference dataset (\wecEng{}) \cite{eirew-etal-2021-wec}. This setup exposes potential limitations of the available annotations in \wecEng{} which might be partially noisy in several manners.

By using Wikipedia as the knowledge source, we assume that the corpus is comprised of high quality documents. Yet, future  work may further assess the quality of the documents inside \wecEng{}, such as checking for duplications.

Second, since the \wecEng{} train set was built using automatic annotation, it might contain some wrong coreference annotations. Wikipedia instructs authors to mark the first occurrence of a mention in the article. However, for several rare occasions, such distracting passages might contain events which were not covered either due to an author not following the instructions or the existence of more than one mentions of the same event within the same passage (§\ref{sec:e2e-results}). While we observed that false-negative retrievals are quite rare, this aspect may be further investigated.

Finally, our dataset covers events which are ``famous" to a certain extent, justifying a Wikipedia entry, but does not cover anecdotal events that may arise in various realistic use cases.

\section*{Acknowledgments}
We thank the Deepset team for providing and supporting the Haystack framework. This research was supported in part by Intel Labs, the Israel Science Foundation grant 2827/21, by a grant from the Israel Ministry of Science and Technology and by the PBC Fellowship for outstanding data science students.

\bibliography{wec_es}
\bibliographystyle{acl_natbib}

\appendix

\clearpage
\appendix

\section{Appendices}

\subsection{Sparse Passage Retriever}
\label{apd:sparse-retriever}
We created a BM25 baseline model following common practice of comparing a retriever model with traditional sparse vector space methods such as BM25 \cite{karpukhin-etal-2020-dense, Khattab2020ColBERTEA}. Additionally, our training procedure depends on challenging negative examples provided by a BM25 model (§\ref{sec:retriever-model}).

In our task settings, a query is represented by a context with mention, to that end, we experiment using different query configurations in order to maximize our BM25 results. This included; using the entire query context, the query sentence, decontextualization \cite{choi-etal-2021} based on the sentence containing the event mention, and using the mention span followed by the Named Entities\footnote{Using spaCy \cite{spacy2} NER} from the surrounding context. We found the latter to gave us the best BM25 results (\emph{BM25} in Table~\ref{tab:ret-results}).

\subsection{A Sample of Erroneous Top Ranked Results}
\label{apd:error}
\begin{table*}
    \centering
    \resizebox{1\textwidth}{!}{
    \begin{tabular}{@{}c|c|c|c@{}}
    \toprule
    \makecell{\textbf{No}} & \makecell{\textbf{Query Context}} &    \makecell{\textbf{Top Result}} & \makecell{\textbf{Error Type}}  \\
    \toprule
        1 & 
        \begin{tabular}{@{}p{10cm}@{}} 
            \textcolor{blue}{On March 4, 2001}, while on his way to his office, Dean was critically wounded in a \textbf{\textcolor{greenrgb}{Palestinian suicide attack}} which took place in the \textcolor{blue}{centre of Netanya}. Dean was rushed to the Hillel Yaffe Medical Center and died five hours later from his wounds. His daughter's sister, Shlomit Ziv, whom he met by chance right before the attack took place, was killed instantly in the attack. Dean was buried in the Tel Mond cemetery. After his death, the first Council house in Tel Mond was named after him - "Naphtali Building"...
        \end{tabular}
        & 
        \begin{tabular}{@{}p{10cm}@{}}
            Speaking before the United Nations Security Council on 24 June 2017, Israeli ambassador Danny Danon, together with Oran Almog, one of the victims of the \textbf{\textcolor{greenrgb}{Maxim restaurant suicide bombing}}, demanded that the \textcolor{blue}{PA} cease incentivizing terrorism by paying stipends to terrorists
        \end{tabular}
        & 
        \makecell{Cannot be \\ determined \\ (Wikipedia)}
        \\
        \midrule
        2 & 
        \begin{tabular}{@{}p{10cm}@{}} 
            ...However, the nascent Republican Party's \textbf{\textcolor{greenrgb}{first convention}} took place in Philadelphia, and the \textcolor{blue}{1860} elections saw the Republican Party win the state's presidential vote and the governor's office. After the failure of the Crittenden Compromise, the secession of the South, and the Battle of Fort Sumter, the \textcolor{blue}{Civil War} began with Pennsylvania as a key member of the Union. Despite the Republican victory the 1860 election, Democrats remained powerful in the state, and several "copperheads" called for peace during the war. The Democrats re-took control of the state legislature in the 1862 election, but incumbent Republican Governor Andrew Curtin retained control of the governorship in 1863. In the 1864 election, \textcolor{blue}{President Lincoln} narrowly defeated Pennsylvania native George B. McClellan for the state's electoral votes
        \end{tabular}
        & 
        \begin{tabular}{@{}p{10cm}@{}}
            Howe was elected as a Whig to the Thirty-second and Thirty-third Congresses. He was not a candidate for renomination in 1854. He resumed his former business pursuits, and was a delegate to the \textbf{\textcolor{greenrgb}{1860 Republican National Convention}} that nominated \textcolor{blue}{Abraham Lincoln} as the candidate for president. He was assistant adjutant general on the staff of Governor Andrew Gregg Curtin and chairman of the Allegheny County committee for recruiting Union soldiers during the \textcolor{blue}{American Civil War}. He was one of the organizers and first president of the Pittsburgh chamber of commerce. He died in Pittsburgh in 1877 and was interred in Allegheny Cemetery
        \end{tabular}
        & 
        \makecell{Cannot be \\ determined \\ (Cluster / Wikipedia)}
        \\
        \midrule
        3 & 
        \begin{tabular}{@{}p{10cm}@{}} 
            A colorless version of the logo is particularly used on a local homepage in recognition of a major tragedy, often for several days. The design was apparently first used on the Google Poland homepage following the air disaster that killed, among others, Polish President Lech Kaczyński in \textcolor{blue}{April 2010}. \textcolor{blue}{A few days later}, the logo was used in China and Hong Kong to pay respects to the victims of the \textcolor{blue}{Qinghai} \textbf{\textcolor{greenrgb}{earthquake}}
        \end{tabular}
        & 
        \begin{tabular}{@{}p{10cm}@{}} 
            She donated her prize money from the tournament and spent time helping the victims and post-reconstruction effort of the \textbf{\textcolor{greenrgb}{12 May earthquake}} that killed nearly 70,000 people and left five to ten million homeless in her home \textcolor{blue}{province Sichuan}. She did the same with her French Open prize money earlier in the year
        \end{tabular}
        & 
        \makecell{Time and location \\ mismatch}
        \\
        \midrule
        4 & 
        \begin{tabular}{@{}p{10cm}@{}} 
            Swift was named ''Billboard''s Woman of the Year \textcolor{blue}{in 2014}, becoming the first artist to win the award twice. \textcolor{blue}{Also that year}, she received the Dick Clark Award for Excellence at the \textbf{\textcolor{greenrgb}{American Music Awards}}. In 2015, "Shake It Off" was nominated for three Grammy Awards, including Record of the Year and Song of the Year and Swift won the Brit Award for International Female Solo Artist...
        \end{tabular}
        & 
        \begin{tabular}{@{}p{10cm}@{}} 
            Bieber performed the song on ''The Ellen DeGeneres Show'' on November  13, 2015. He was also a musical guest on ''The Tonight Show Starring Jimmy Fallon''. Additionally, Bieber performed the song during the \textbf{\textcolor{greenrgb}{2015 American Music Awards}}, which took place at Microsoft Theater on 22 November 2015 in Los Angeles, California. The singer also took the stage to perform "Sorry"...
        \end{tabular}
        & 
        \makecell{Year mismatch}
        \\
        \midrule
        5 & 
        \begin{tabular}{@{}p{10cm}@{}} 
            Johannes Barge (23 March 1906 – 28 February 2000) was an officer in the Wehrmacht of Nazi Germany during World War II who was responsible for German military operations causing \textbf{\textcolor{greenrgb}{Cephalonia Massacre}} in \textcolor{blue}{September 1943}
        \end{tabular}
        & 
        \begin{tabular}{@{}p{10cm}@{}} 
            ...On 18 June 1944, EDES forces with Allied support launched an attack on \textcolor{blue}{Paramythia}. After short-term conflict against a combined Cham-German garrison, the town was finally under Allied command. Soon after, violent reprisals were carried out against \textcolor{blue}{the town's} Muslim community, which was considered responsible for the \textbf{\textcolor{greenrgb}{massacre of September 1943}}
        \end{tabular}
        & 
        \makecell{Location mismatch}
        \\
        \midrule
        6 & 
        \begin{tabular}{@{}p{10cm}@{}} 
            The region had previously experienced one of the worst earthquakes in 1897, measuring 8.1 on the Richter scale, that claimed the lives of over 1,500 people. Again in \textcolor{blue}{September 2011}, more than \textcolor{blue}{50 people died} after a \textbf{\textcolor{greenrgb}{killer quake measuring 6.9}} had shook the region
        \end{tabular}
        & 
        \begin{tabular}{@{}p{10cm}@{}} 
            The \textcolor{blue}{7.2} \textbf{\textcolor{greenrgb}{Dalbandin earthquake}} shook a remote region of \textcolor{blue}{Balochistan} on \textcolor{blue}{19 January 2011}. The dip-slip shock had a maximum Mercalli intensity of VI (''Strong''), caused moderate damage, and left \textcolor{blue}{three dead} and several injured
        \end{tabular}
        & 
        \makecell{Several mismatched \\ entities}
        \\
    \bottomrule
    \end{tabular}}
    \caption{A sample of queries with top result marked as false (i.e., containing an event not coreferring with the query event), produced by the E2E-Integrated model. \textbf{\textcolor{greenrgb}{Green}} signifies an event mention span. \textcolor{blue}{Blue} represents some of the event arguments (such as time, location, participants, etc.) that may indicate whether the query and result events hold a coreference relation or not.}
    \label{tab:err-analysis}
\end{table*}


\end{document}